# Heart disease risk prediction using deep learning techniques with feature augmentation

María Teresa García-Ordás[1] · Martín Bayón-Gutiérrez[1] · Carmen Benavides[2] · Jose Aveleira-Mata[1] · José Alberto Benítez-Andrades[2]

## Abstract
Cardiovascular diseases state as one of the greatest risks of death for the general population. Late detection in heart diseases highly conditions the chances of survival for patients. Age, sex, cholesterol level, sugar level, heart rate, among other factors, are known to have an influence on life-threatening heart problems, but, due to the high amount of variables, it is often difficult for an expert to evaluate each patient taking this information into account. In this manuscript, the authors propose using deep learning methods, combined with feature augmentation techniques for evaluating whether patients are at risk of suffering cardiovascular disease. The results of the proposed methods outperform other state of the art methods by 4.4%, leading to a precision of a 90%, which presents a significant improvement, even more so when it comes to an affliction that affects a large population.

María Teresa García-Ordías, Martín Bayón-Gutiérrez, Carmen Benavides, Jose Aveleira-Mata and José Alberto Ben´ıtez-Andrades are contributed equally to this work.

✉ José Alberto Ben´ıtez-Andrades
  jbena@unileon.es

  María Teresa García-Ordás
  mgaro@unileon.es

  Martín Bayón-Gutiérrez
  martin.bayon@unileon.es

  Carmen Benavides
  carmen.benavides@unileon.es

  Jose Aveleira-Mata
  jose.aveleira@unileon.es

1 SECOMUCI Research Group, Escuela de Ingenierías Industrial e Informática,
  Universidad de León, Campus of Vegazana s/n, León, 24071, León, Spain

2 SALBIS Research Group, Department of Electric, Systems and Automatics Engineering,
  Universidad de León, Campus of Vegazana s/n, León, 24071, León, Spain

## 1 Introduction and related work

Cardiovascular diseases (CVDs) are the main reasons for disease burden and mortality all over the world [22]. The term "cardiovascular disease" includes a wide range of conditions affecting the heart and blood vessels and the way blood is pumped and circulated through the body [29]. Heart disease is a common disease which has given rise to the deaths of many people. This is because it affects heart function and may cause death [9]. Over the last few decades, the population worldwide is has increasingly suffered from heart disease,considered one of the most significant causes of fatalities. About 17.7 million people die anually because of heart disease [19]. Diagnoses related to heart disease are made by a specialized doctor and it is essential for good treatment. This has the disadvantage that diagnoses may not be entirely objective and is subject to human error.

Heart failure has been subject to significant research as a result of its complicated diagnostic procedure [15] making a Computer Aided Decision Support System very helpful in this field, as the one presented in [20], where data mining techniques were used to reduce the time it takes to make an accurate prediction of the disease

Heart diseases are highly varied and lead to different types of complications that can lead to reduced quality of life and even death, especially in developing countries [30]. Furthermore, the number of deaths that occurs due to heart failure is higher in developing countries and in those with worse health facilities [23]. This highlights the need for the development of a method that can guarantee an accurate and early prediction of the risk of heart failure in patients.

For these reasons, many authors have developed methods that help in the detection of heart disease, by taking into account different factors. Most of this methods use machine learning techniques to prevent the problems derived from statistical analysis methods, that fail to capture prognostic information in large datasets containing multi-dimensional interactions [3, 17, 18, 25, 26]. Some of these papers have generally benefited from large datasets that allow detection of existing diseases thanks to historical data over a long period of time. On the other hand, the results that were obtained until a few years ago, generally focused more on determining abnormal behaviors of the heart without going into detail about whether they are possible serious cases or simply benign to health.

In [1], the authors used a boosted decision tree algorithm to capture correlations between patient characteristics and mortality. Patients were classified as having a high or low risk of death based on their early death or not of them after visiting the hospital. Results obtained an area under the curve of 0.88 considering the proposed labels.

Pires et al. proposed in [27] the use of different methods, such as Neural Networks, Decision Trees, k-Nearest Neighbor (kNN), Combined nomenclature (CN2) rule inducer, Support Vector Machine (SVM), and Stochastic Gradient Descent (SGD) obtaining results of up to 87.69% which is a good result in the case of predicting heart disease. The article proposes a method that is capable of detecting heart disease with an acceptable success rate, but validations have been carried out with a limited number of individuals.

More recently, Ali et al. [4] tried to identify the best machine learning classifiers -that with the highest accuracy- for these diagnostic purposes. Several supervised machine-learning algorithms were applied and compared for performance and accuracy in heart disease prediction. Results showed a 100% of accuracy using Random Forest, Decision Trees and kNN but they presented only the best result achieved after a cross-validation process which is not very conclusive.

Those papers used classic methods to predict heart diseases. However, the precision of the classifications can be enhanced by using improved techniques, as in the case of Faiayaz et al., [11], which managed to increase the accuracy by 5.68% compared to the original kNN, by proposing an improved KNN. In [19] a hybrid method of Random Forest with a Linear Model (HRFLM) technique reached an 88.7% precision on the Cleveland dataset which contains 297 records.

All these works seem very promising but lack the need to generalize with a larger number of patients.

As classical machine learning models showed promising results for this problem, many other authors have tried to combine different machine learning algorithms.

For example, in [24], many machine learning techniques, such as generalized boosted regression, main-terms Bayesian logistic regression using a Cauchy prior, penalized regression, main-terms logistic regression with and without variable selection, bootstrap aggregation of regression trees, multivariate adaptive regression splines, the arithmetic mean assigning the marginal probability of mortality to each patient, and classification and regression trees using random forest, were used in a 30-day mortality risk prediction after discharge in patients with heart failure. Results showed that ensemble models achieved better results than the benchmark models.

Authors in [16] combined five classifier model approaches, including support vector machine, artificial neural network, Naïve Bayes, regression analysis, and random forest, to predict and diagnose the recurrence of cardiovascular disease. In this case, Cleveland and Hungarian datasets from the UCI repository were used. Results demonstrated the better performance of ensemble algorithms obtaining the best results with the random forest method (98.12% of accuracy).

Ensemble algorithms have demonstrated well performance not only in heart failure detection but also in another diseases such as breast cancer [8], Hepatitis C [10] or Diabetics retinopathy [14].

In the last years, deep learning techniques based on neural networks have also been used to address medical problems including heart failure risk.

Convolutional Neural Networks (CNN) was used in [2] for the early identification of individuals at risk of heart failure using solely electrocardiograms (ECGs) obtaining a 0.78 AUC. Adaptive multi-layer networks were also used in [28] to predict the risk of heart failure, outperforming classical neural networks and even hybrid and ensemble techniques proposed in the previous years. In this work, the Cleveland dataset was used, so the number of samples was small with only 297 patients evaluated.

In the last year, a new dataset [12] consisting of some well-known datasets such as Cleveland (303 observations), Hungary (294 observations), Switzerland (123 observations), Long Beach VA (200 observations), and Stalog (270 observations) allowed the training of new techniques, that were capable of classifying this high volume of samples with a very limited number of features (11).

The main goal of this paper is:

- Obtaining results for the classification of heart diseases that allow us to have a high percentage of successes in the early detection of the disease.

Taking into account the dataset evaluated, two secondary goals have been achieved:

- Finding a new methodology for classification problems with a very low number of features.

- Harnessing the properties of convolutional neural networks to improve current feature augmentation techniques.

To achieve these goals, in this work, an architecture based on convolutional neural networks and a sparse autoencoder is proposed for the treatment of the data and its subsequent analysis, obtaining precisions of up to 90% , which can be considered a significant advance and a great help in determining the risk of heart diseases.

The rest of this paper is organized as follows: In Section 2, the classic methods used to compare results and the architectures used in our proposal are explained. Experiments and results, as well as the dataset used, are detailed in Section 3. Finally, we conclude in Section 4.

## 2 Methodology

### 2.1 Classic methods

Several classic methods have been used to compare the results with the proposed architecture. They are going to be introduced very briefly as they are all well-known methods.

#### 2.1.1 Decision tree

A decision tree is a model in which each internal or intermediate node is labeled with an input feature. These intermediate nodes are those in which a decision must be made between several possible ones. Arcs are the unions between nodes and come from a node labeled with one input feature which is labeled with each of the possible values of the output or target feature. Alternatively, the arc leads to a subordinate decision node on a different input feature. Leaf nodes are labeled with a class or probability distribution between classes, which means that the tree has classified the data set into a specific class or a particular probability distribution so these nodes contain the final selected class.

#### 2.1.2 Random forest

Random forest [6] is a combination of decision trees. Each of the decision trees that make up the forest is built as follows: First, the number of data (N) and the number of variables of the classifier (M) are defined. The number of input variables that is used to determine the decision of a certain node is called $m$. For each node of the tree, $m$ variables are chosen and from these $m$ variables, the best partition of the set is calculated. To predict a new case, the nodes of the tree are traversed downwards and the label of the terminal node it reaches is assigned to it. This process is iterated throughout all the trees in the forest and the one that obtains the highest number of indices will be the one used as a predictor. Random forest is one of the most accurate learning algorithms, as long as a large enough data set is used [7].

#### 2.1.3 K-nearest neighbors

The k nearest neighbors method is one of the simplest. Unlike other machine learning algorithms, k-NN does not generate a model from the training data, but rather the learning takes place at the same moment in which it is tested with the test data, therefore, it is a lazy learning method. The input data are vectors of dimension p of the form:

$$x_i = (x_{1i}, x_{2i}, \ldots, x_{pi}) \in X \quad (1)$$

In the training phase, the vectors and class labels of the training examples are stored and the distance between the stored vectors and the new vector is calculated in the classification phase, and the $k$ examples closest to that new input data are selected. The new data are classified with the class that repeats the most in the selected vectors. Any metric can be used to calculate the distance, but the most common one is to use the Euclidean distance (See (2)).

$$d(x_i, x_j) = \sqrt{\sum_{r=1}^{p} (x_{ri} - x_{rj})^2} \quad (2)$$

### 2.1.4 AdaBoost

AdaBoost consists of combining several weak classifiers in order to obtain a robust classifier. The main idea is to assign greater weights to poorly classified data and to assign less weight to data that has been well classified. Thus, each weak classifier focuses more on badly classified cases, thus improving the results. In our case, decision trees have been used as the base method. The algorithm is made up of three steps. First, the weights are initialized and each of the $N$ samples is given the same weight. In a second step, a weak classifier is trained taking into account that if the data is correctly classified the weight is reduced and if it is badly classified, the weight is increased to give it more importance. Finally, the weak classifiers obtained in each training are combined into a strong classifier.

### 2.1.5 XGBoost

The XGBoost classifier has also been used to compare the results obtained with those obtained with our proposal. The difference between this method and AdaBoost is that in each iteration, instead of assigning more weight to misclassified samples, XGBoost focuses on reducing losses. Each iteration focuses on reducing the error and establishing a new model to reduce the loss (negative gradient) further.

### 2.1.6 Gaussian Naive Bayes (GNB)

This classifier is based on Bayes' theorem. The predictor variables are assumed to be independent of each other. First, the data set is converted into a frequency table. A probability table is created, Bayes' theorem is applied and the class with the highest posterior probability is the result of the prediction. It works in the same way as the Naive Bayes classifier but in this case GNB follows a Gaussian normal distribution and supports continuous data.

### 2.1.7 Multilayer Perceptron (MLP)

A multilayer perceptron is a neural network that has an input layer, an output layer and one or more intermediate layers with a certain number of neurons. It has the peculiarity that it has a linear activation function in all neurons and each neuron of a layer is connected with the neurons of the previous and next layer learning complex information on the input data.

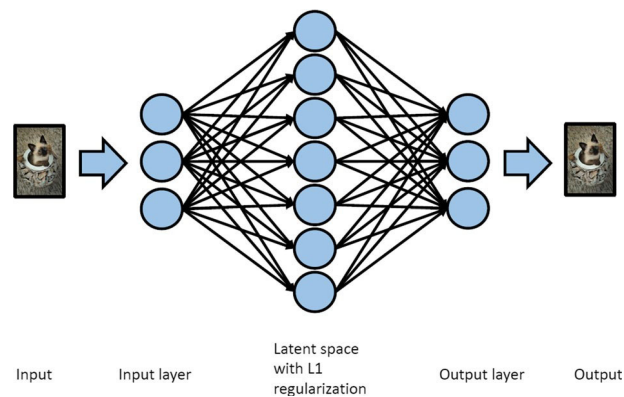

**Fig. 1** Vanilla SAE network architecture. In the latent space, and L1 regularization term is applied

### 2.2 Proposal

#### 2.2.1 Reconstruction and feature augmentation: Sparse Autoencoder (SAE)

In a conventional autoencoder, the latent space has fewer neurons than the input and output layers, so it is possible to represent a feature vector in a reduced way. In the case of the sparse autoencoder (SAE), the latent space has more neurons than the input and output layers. An L1 regularization term is also added to the latent space to force the network to just use some neurons each time. With this type of network we manage to increase the number of features of the data and analyze them from a different perspective.

In Fig. 1 the typical architecture of a SAE can be seen.

Once the Sparse Autoencoder has been trained for input reconstruction, the decoder part of the neural network can be detached, leaving only the part of the encoder that reaches the latent space. The encoder increases the features of the input data in such a way that the initial N features are dissociated to form M features (with N<M). This procedure allows to keep all the information of the original data and add additional information that was hidden.

#### 2.2.2 Data classification: Convolutional Neural Networks (CNN)

Convolutional neural networks can take two-dimensional data, such as images, as input and are capable of extracting complex features from that data. This information is extracted thanks to the filters (kernels). In the training process, the filter weights are adjusted to carry out an accurate feature map of each class.

In the architecture of a CNN, each convolutional layer must be followed by a pooling layer. The pooling layers are responsible for reducing the overfitting and the number of network parameters so that the computation is not so heavy. The most widely used type of pooling is Max-pooling, which acts by selecting the maximum value of each window. The last layers of the convolutional network have to be dense layers to be able to carry out the classification of the features extracted by the kernels. A vanilla CNN is represented in Fig. 2

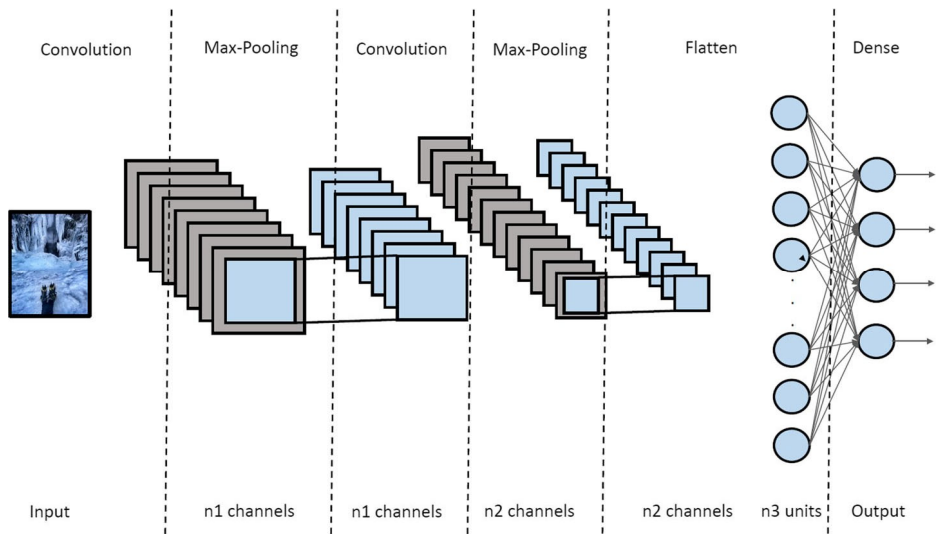

**Fig. 2** A vanilla CNN representation

### 2.2.3 Our proposal: multi task neural network

The proposed architecture consists of two tasks. On the one hand, the features of each data are augmented using a sparse autoencoder (SAE). Furthermore, the latent space is connected with a classifier which is trained together with the SAE. We have evaluated two different classifiers (to join with the SAE): a traditional MLP (see Fig. 3) and a convolutional neural network (see Fig. 4). The two tasks are described independently below.

In Fig. 5, three different approaches dealing with data with a small number of features are presented. The first approach takes as an input the whole training dataset and carries out a feature augmentation process to increase the number of them. This new dataset with more features than the original one is took as the input of the classification model, determining if the sample has a positive class or not.

Second approach ignores the feature augmentation step and directly feed the model with the small dataset.

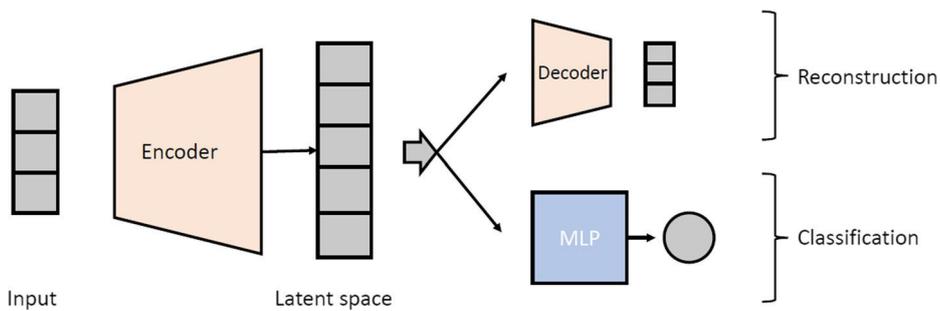

**Fig. 3** Multitask neural network composed of Sparse Autoencoder and MLP classifier

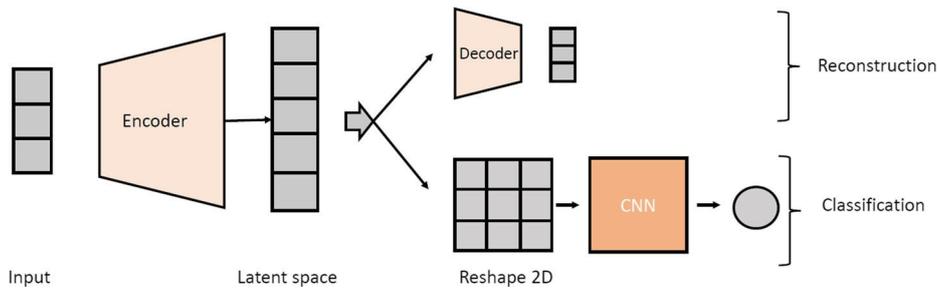

**Fig. 4** Multitask neural network composed of Sparse Autoencoder and CNN classifier

Our proposed approach is based on a parallel training process of a data augmentation neural network called Sparse Autoencoder and a classifier network which takes as the input the maximized dataset to predict the final class. Training a model with this approach makes the process of augmenting features something trainable, and as information is processed, it is capable of improving to achieve more accurate predictions.

## 3 Experiments and results

### 3.1 Dataset

The dataset used is made up of 11 clinical features: the patient's age, sex, type of chest pain (typical angina, atypical angina, non-anginal pain or asymptomatic), the resting blood pressure mmHg, the serum cholesterol (mm/dl), the fasting blood sugar (value 1 if FastingBS > 120 mg/dl, and value 0 otherwise), resting electrocardiogram results (which can be Normal, ST if the patient has ST-T abnormalities or LVH if the patient shows probable ventricular hypertrophy), the maximum heart rate (numeric value between 60 and 202), exercise-induced angina which can be yes or no, the oldpeak (numeric value measured in depression) and finally, the slope of the peak exercise ST segment (Up, Flat, Down). The column number 12 contains the output class which can be 1 (heart disease) or 0 (normal).

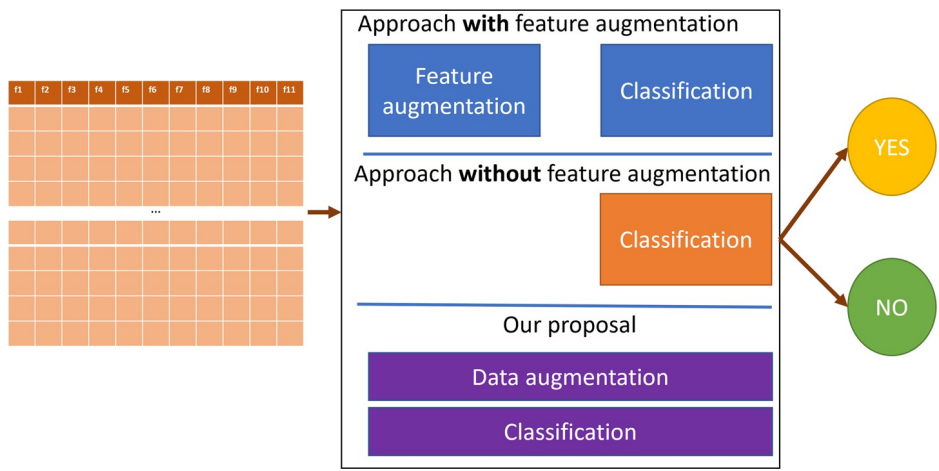

**Fig. 5** Schema of three different approaches dealing with data with a small number of features

This dataset was created on September 2021 by combining different datasets already available independently, but not combined before: Cleveland, Hungarian, Switzerland, Long Beach, stalog. The final heart disease dataset is made up of 918 samples with a similar number of cases for each class, 410 corresponding to the healthy class and 508 to the with heart issues class [12]. For that reason, no methods have been needed to deal with unbalanced classes.

### 3.2 Experimental setup

First, a preprocessing step was carried out in order to clean and extract more useful information from the dataset. The age was dropped and three new columns were added representing an age range: young, adult and elder. In the same way, the feature resting BP was converted into three new columns for lowBP, mediumBP and highBP. Lastly, the Cholesterol feature was converted into three categorical columns which determines the risk: low, medium, high.

One hot encoding technique was applied on three features: ChestPainType, RestingECG and ST_Slope. Finally, Sex and Exercise Angina were processed using a label encoder. At the end of this process, we have a dataset made up of 24 features.

A k-fold cross-validation has been carried out with 10 folds for all of the experiments in order to avoid randomness. Every model has been evaluated through an extensive hyperparameter grid search. In the results, the score for the hyperparameter configuration with the best mean value of the 10 folds is presented. Neural-network architectures are comprised of one sparse autoencoder, for feature augmentation, and one classifier both of which are trained at the same time. We have carried out two different configurations. The first one uses an MLP classifier and the second one converts the latent space into a bidimensional matrix to train a 2D convolutional neural network. In both cases, the training was carried out using the ADAM optimizer due to its fast optimization time, the invariance to rescaling of the gradient and its possibility of working with sparse gradients which highly increase the performance of the neural network.

The selected loss function was binary cross-entropy for the classification subnet and a mean squared error for the decoder. Binary cross-entropy can be defined as:

$$-(y \log(p) + (1-y) \log(1-p)) \qquad (3)$$

and it is used because it is equivalent to fitting the model using maximum likelihood. Mean Squared error can be defined as:

$$\sum_{i=1}^{D} (x_i - y_i)^2 \qquad (4)$$

and it is used because it penalizes a lot the large error which is very interesting for feature reconstruction in the autoencoder problem.

Different latent space sizes were evaluated to study the importance of this parameter in the final classification.

### 3.3 Results

To compare the proposed method along other approaches, authors have carried out an analysis of classical machine learning methods discussed in Section 2.1. As discussed, a grid search has been applied to find the best hyperparameters for each method. This results can be observed in Fig. 6, where the mean of the 10-fold validation with the best configuration is presented.

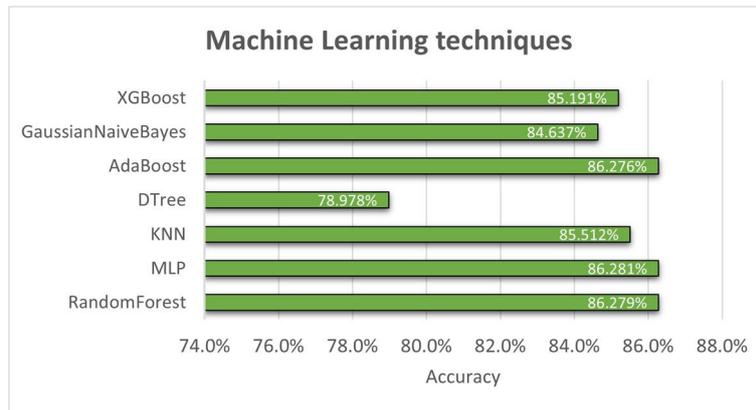

**Fig. 6** Results obtained using classical machine learning methods

These results demonstrate the good performance of the MLP neural network, which achieved the best accuracy with 86.281%. Following that, the Random forest or the Adaboost ensemble method also performed similarly. In contrast, decision trees obtained the worst performance with a 78.978% of accuracy, 8.46% lower than MLP.

Because of the good performance of neural networks, we have carried out training by combining MLP with feature augmentation through an sparse autoencoder. In Fig. 7 we present see the results achieved with different latent space sizes. All of the values represent the mean accuracy on a 10-fold cross-validation.

When the latent space has a size of 100 features, in contrast to the original 11 features, the accuracy of the classification improved by 3.78% achieving an accuracy of 89.543%.

As preliminary results exposed improvements in the classification when using a multi-tasking neural network for training the SAE and the MLP at the same time and given that this approach not only performs the best for heart problem detection but it also allows to the generation of new features from the data set, a new set of experiments have been carried out by replacing the MLP classifier with a 2D CNN. This approach helps CNN to deal with structured data by rearranging its data using SAE adding the optimal spatial representation by reordering the features and creating new ones as a combination of them. In Fig. 8 results with different latent space sizes can be compared.

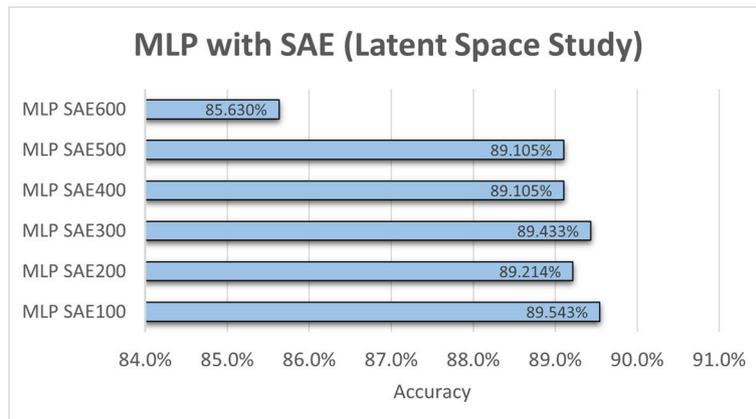

**Fig. 7** Results obtained using the architecture which combines MLP for classification and sparse autoencoder for feature augmentation

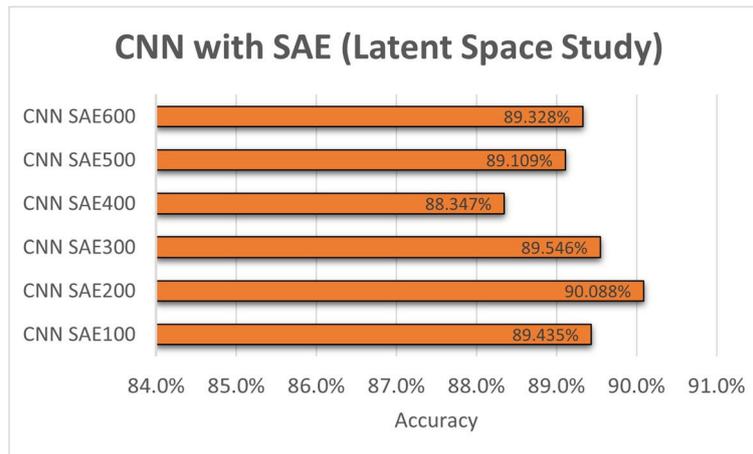

**Fig. 8** Results obtained using the architecture which combines CNN for classification and sparse autoencoder for feature augmentation

In this new set of experiments, the best result was achieved with a latent space of 200 new features with an accuracy of 90.088% increasing the performance over the classic MLP by more than 4.4% which is a really interesting increase, even more so when it comes to a problem that can cause serious problems in patients, even death.

In Fig. 9, we can see the improvement in our proposals over the vanilla MLP and the random forest model. As indicated, feature augmentation techniques lead to the improve the

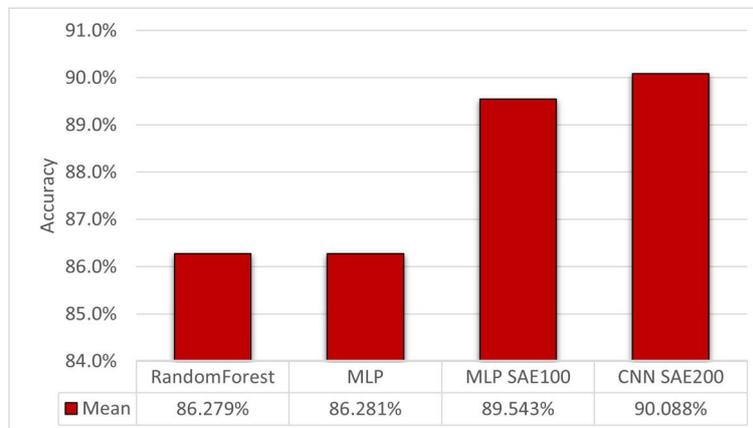

**Fig. 9** Comparison of our proposal multi task neural networks with the classical MLP and Random Forest models

**Table 1** Comparison with results obtained in the state of the art

| Method | Accuracy (%) |
|---|---|
| Our proposal | **90.09** |
| Adaptive Boosting [5] | 85.28 |
| Bagging [5] | 86.37 |
| Stacking [5] | 87.24 |
| RandomForest [13] | 86.40 |
| CatBoost [21] | 89.42 |
| Stacking [21] | 89.86 |

The bold entry is the most accurate model results achieved in the original dataset. Furthermore, the approach of combining SAE with a 2D Convolutional Neural Network to rearrange the new features increases the accuracy slightly.

Two statistical tests were performed to verify that the results obtained by the proposed approach are superior to those obtained by more traditional techniques: the Kolmogorov-Smirnov test and the Independent Samples t-test.

To perform these tests, the accuracy results were grouped as follows: (group I) existing techniques, where traditional machine learning and traditional MLP techniques were included, and (group II) techniques proposed in this paper, i.e. MLP and CNN using SAE. The results after applying these tests exposed that the accuracy superiority obtained by the proposed approach (M=88.99, SD=1.13) is statistically significant in relation to the accuracy obtained using traditional methods (M=84.73, SD=2.61) with a significance of $p<0.001$. In the case of the t-test, the t(17)=4.97, $p<0.001$.

Some recent works have also take advantage of the dataset used in this research. In [5], a study with different machine learning methods was performed. An stacking technique formed by K-NN and Logistic Regresion followed by a k-NN classification using the vote classifier output of the previous techniques obtained the best results with a 87.24% of accuracy. In 2022, Ghosh et al. [13] achieved an accuracy of 86.40% when using Random Forest Model Lately, in [21] an stacking algorithm was proposed which combines the output of Logistic Regression, Random Forest, MLP, Cat Boost and Decision Trees and trains a meta classifier to derive the final result. This method achieved an accuracy of 89.86% over the same experimental conditions than our proposal.

In Table 1 a comparison with the state of the art results and our proposal is presented. Our results using the multitasking classifier with CNN outperforms all the other published methods.

## 4 Conclusions

This article provides deep learning-based methods that allow the combination classification and feature augmentation tasks to address the prediction of heart problems in a dataset consisting of patient records from five independent centers. This dataset consists of 918 samples with only 11 clinical characteristics per sample. A new architectural approach has been proposed that combines the Sparse Autoencoder and the Convolutional Classifier.

As the dataset only contains 11 features, a feature augmentation has been carried out using the Sparse autoencoder to extract new features. Thanks to the high number of features we have extracted, a convolutional neural network can be trained by reshaping them into a 2D array. These two processes are joint in a complex net, which combines SAE and the classifier (MLP or CNN), that has been implemented in order to increase the feature extraction ability by taking into account the classifier information obtained as feedback in the backpropagation algorithm. When the SAE is trained jointly, CNN outperforms MLP in a 0.6% of accuracy. It indicates that CNN interferes in the SAE feature extraction by forcing it to extract more relevant features with spatial location information. MLP also modify the feature extraction carried out by SAE but the improvement is clearly less significant than in the convolutional network.

A deep analysis of the number of neurons in the latent space of the sparse autoencoder, which represents the new features, were performed, concluding that the optimal size was 200. This study is very interesting because it demonstrates that there is a certain size from where the results worsen, which implies that not always the more neurons the better

With this approach, we have achieved 90.088% which represents a 4.4% improvement in comparison with the results obtained by classic classifiers (MLP or RF) trained on the same dataset and under the same conditions.

In addition, our method also obtained better results than those that compose the state of the art, that carried out combinations of algorithms (stacking). Moreover, stacking is a computationally very expensive technique, since it involves tha analysis of several models, sometimes sequentially, to obtain the desired result.

Taking into account that detecting a heart problem in a patient can mean survival, improvements presented in this manuscript, along with the proposed method, are of great interest to specialists in the field.

**Author Contributions María Teresa García-Ordás**: Conceptualization, Data curation, Methodology, Software, Visualization, Validation, Writing- Original draft preparation. **Martín Bayón-Gutiérrez**: Data curation, Writing- Original draft preparation.**Carmen Benavides**: Conceptualization, Supervision, Writing- Reviewing and Editing. **Jose Aveleira-Mata**: Conceptualization, Supervision, Writing- Reviewing and Editing. **José Alberto Benítez-Andrades**: Data curation, Methodology, Software, Visualization, Validation, Writing- Reviewing and Editing.

**Funding** Open Access funding provided thanks to the CRUE-CSIC agreement with Springer Nature. This research was funded by the Junta de Castilla y Leon grant number LE078G18. This work is partially supported by Universidad de León under the "Programa Propio de Investigación de la Universidad de León 2021" grant.

**Data Availability** All the data used in the experiments, are available in a Kaggle https://www.kaggle.com/fedesoriano/heart-failure-prediction

### Declarations

**Competing interests** The authors declare that they have no known competing financial interests or personal relationships that could have appeared to influence the work reported in this paper.